\documentclass[runningheads]{llncs}
\usepackage{graphicx}

\usepackage{algorithm}
\usepackage{algpseudocode}

\usepackage{amsthm}
\usepackage{amsmath}
\usepackage{lipsum}
\usepackage{tikz}
\newlength{\indentationFormule}
\newlength{\indentationTotaleFormule}
\newlength{\indentationCommentaire}
\newlength{\indentationDerivation}
\newlength{\largeurLangle}
\newlength{\largeurBoiteCommentaire}

\theoremstyle{definition}

\newenvironment{deriv}[1][\leftmargini]%
{\setlength{\indentationDerivation}{#1}%
	\setlength{\indentationTotaleFormule}{\indentationFormule}
	\addtolength{\indentationTotaleFormule}{#1}
	\setlength{\largeurBoiteCommentaire}{\linewidth}
	\addtolength{\largeurBoiteCommentaire}{-\indentationFormule}
	\addtolength{\largeurBoiteCommentaire}{-\indentationCommentaire}
	\addtolength{\largeurBoiteCommentaire}{-\largeurLangle}
	\addtolength{\largeurBoiteCommentaire}{-\indentationDerivation}
	\begin{list}{}{\setlength{\leftmargin}{\indentationTotaleFormule}}
		\setlength{\baselineskip}{1.3\baselineskip}
		\item$}%
	{\hbox{}$\end{list}}  

\newcommand{\<}[1]{\\\hspace*{-\indentationFormule}\makebox(0,0)[bl]{$#1$}\hspace*{\indentationFormule}}

\newcommand{\commentaire}[1]{\hspace*{\indentationCommentaire}\langle\hspace*{.4em}%
	\begin{minipage}[t]{\largeurBoiteCommentaire}%
		#1 $\rangle$\\\hbox{}
	\end{minipage}\\[-1ex] 
}

\newenvironment{arra}[1]{\begin{array}{@{}#1@{}}}{\end{array}}

\newcommand{\ssi}{\Leftrightarrow}

\newcommand{\imp}{\Rightarrow}

\newcommand{\diff}[1]{-}

\newcommand{\quant}[4]{(#1#2 \mid #3:#4)}

\newcommand{\symbpourtout}{\forall}
\newcommand{\pourtout}[3]{\quant{\symbpourtout}{#1}{#2}{#3}}

\newcommand{\Max}[3]{\quant{\max}{#1}{#2}{#3}}
\newcommand{\Min}[3]{\quant{\min}{#1}{#2}{#3}}

\newcommand{\Union}[3]{\quant{\bigcup}{#1}{#2}{#3}}

\newcommand{\Inter}[3]{\quant{\bigcap}{#1}{#2}{#3}}

\newcommand{\SQCAP}[3]{{\vcenter{\vbox{\hrule height.#3pt%
				\hbox{\vrule width.#3pt height#1pt \kern#2pt \vrule%
					width.#3pt}}}}}

\newcommand{\relcompsymb}{\kern-.5pt\raise.3ex\hbox{\footnotesize;}\kern-.5pt}

\makeatletter
\@ifundefined{smallsmile}{%
	\def\conversesymb{\cup}
}{%
	\def\conversesymb{\smallsmile}
}
\makeatother
\newcommand{\internalconverse}[1]{#1^{%
		\mkern-1mu{}{\raise0.3ex\hbox{\tiny$\conversesymb$}}}%
	\kern-0.1em{}}

\begin{document}
\title{Skeletonization and Reconstruction based on Graph Morphological Transformations}

\titlerunning{Graph Morphological Skeletonization and Reconstruction}
%

\author{Hossein Memarzadeh Sharifipour \inst{1*}\and
Bardia Yousefi \inst{2,3} \and
Xavier P.V. Maldague \inst{2}}
%
%
\authorrunning{H.M. Sharifipour, B. Yousefi \& X. Maldague}
%
\institute{Department of Computer Science, Laval University, Québec, CA \email{Hossein.MemarzadehSharifipour.1@ulaval.ca}\and
Department of Electrical and Computer Engineering, Laval University, Québec, CA
\\\and
\textit{Address:} University of Pennsylvania, Philadelphia PA 19104\\
\email{Bardia.Yousefi.1@ulaval.ca}, \email{Xavier.Maldague@gel.ulaval.ca}}
\maketitle              
\begin{abstract}
Multiscale shape skeletonization on pixel adjacency graphs is an advanced intriguing research subject  in the field of image processing, computer vision and data mining. The previous works in this area almost focused on the graph vertices. We proposed novel structured based graph morphological transformations based on edges opposite to the current node based transformations and used them for deploying   skeletonization and reconstruction of infrared thermal images represented by graphs. The advantage of this method is that many  widely used path based approaches become available within this definition of morphological operations. For instance, we use distance maps and image foresting transform (IFT) as two main path based methods are utilized for computing the skeleton of an image. Moreover,   In addition, the open question proposed by Maragos et al. about connectivity of graph skeletonization method are discussed and shown to be quite difficult to decide in general case.

\end{abstract}

\section{Introduction}

Mathematical morphology is a highly used framework for binary and grayscale image processing \cite{heijmans1990algebraic,x1} (also for infrared image analyses \cite{x2,x3,x4,x5,x6}). The core of mathematical morphology depends upon fitting a pattern called the structuring element on an image for extracting useful information about the geometrical structure of the image by matching small patterns into it at various locations of the image. By making use of different shapes and sizes of structuring elements, information such as connectivity and  borders considering the shape of different parts of the image and interrelations between them can be obtained.

Graph based image representations typically deal with pixel adjacency graphs, i.e, graphs whose nodes are the set of image pixels and whose edge set is determined by some adjacency relation on image pixels. In a 2D grid environment the 4,6,8-connected adjacency relation applies according to the connectivity property of the grid. Figure 1 indicates graph representation of the two dimensional grid with respect to 4,8-connectivity. 

Extensive applications of graphs for clustering, segmentation and edge detection in different areas of image processing and infrared thermography \cite{yousefi2014hierarchical},\cite{yan2015infrared} \cite{vollmer2017infrared},\cite{elibol2014graph}  have been proposed recently. 

Exposing a gray-level image by a graph express the pixel values of the image by a function  from the set of vertices $V$ to the complete lattice of gray values $f:V \to L$. In binary case where pixels are only black or white, this complete lattice just contains two elements $\{0,1\}$.

 The notion of an image skeleton has been introduced by Bium \cite{biumtransformation}. He called his idea the "medial axis" and later the "symmetric axis".
The intuitive model suggests considering an image as a grass field. At time t=0 the image boundary is set to fire and the fire propagates inwards \cite{beucher1994digital}. The set of points remaining after extinguishing the fire is adopted as the skeleton of the image. An applicative and very useful property of the symmetric axis is inherited in its capability of reconstructing the image boundary by propagating the wavefronts backward.
 Lantuéjoul employed binary morphological operations for calculating the skeleton of a binary image \cite{lantuejoul1978squelettisation}. He proved that the skeleton Skel(M) of the image M can be extracted by an iterative routine as follows,
\begin{equation}  
\mathsf{skel}_n(M)=(M \ominus nB)-((M \ominus nB) \circ \mathsf{cl}B)
\end{equation}
where $nB$ denotes dilation of the structuring element $B$ with itself iterated for $n$ times and $\ominus,\circ$ and $clB$  correspond to the binary erosion, opening and closed structuring element $B$ respectively. This leads to an iterative approach for calculating the skeleton. The iteration continues by $n$ steps until $M \ominus nB \neq \emptyset $. Kresh \cite{kresch1994morphological} applied morphological operations for extracting skeletons while Maragos et al. have utilized the method for encoding binary images \cite{maragos1986morphological}.

 Given the skeleton of an image, the original image can be completely or partially retrieved by the following formula  called reconstruction,
 \begin{equation}
 M \circ nB =\Union{n} { k \leq n \leq N}{\mathsf{skel}_n(M)\oplus nB} 
 \end{equation}
 where $k \geq 0$  is an integer controlling the process and $\oplus, \circ$ are  dilation and opening operations respectively.  By setting $k=0$, one gets a complete reconstruction of the original image whereas $k \geq 1$ conducts partial reconstruction of original image. The result of reconstruction in this case corresponds to a copy of the opened original  image by the structuring element $nB$.

 \begin{figure*}[!t]
	\centering
\includegraphics[scale=0.38]{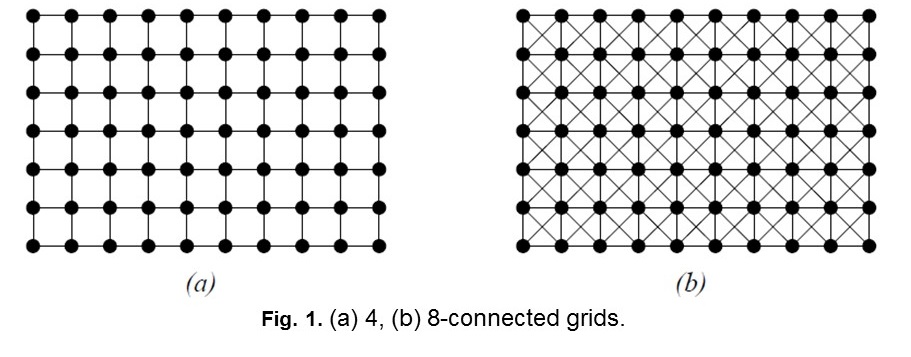}
	\label{fig:1}
\end{figure*}
    Exploiting binary morphological operations introduced in \cite{heijmans1990algebraic} to graph morphological transformations were first proposed by Maragos \cite{maragos2013segmentation,maragos1990morphological}.  We have combined the idea of structured binary morphological operations proposed initially in \cite{vincent1989graphs} and edge based morphological operations explained  in \cite{najman2012short} to attain new morphological operations for graphs.  Then the skeletonization procedure is formalized benefiting from these new binary morphological operations. Comparing with Maragos technique that uses vertex based morphological operations our method opens the door to import many well-known path related methods such as distance mappings into graph based morphological transformations. Moreover, the proposed method was employed for extracting skeleton of some infrared thermal samples.
Before proposing our method,  we intend to respond the question posed by Maragos in \cite{maragos2013segmentation} about  connectivity of skeletonizing method using graph. The more general question that can be discussed is under which conditions morphological transformations defined for graphs are connected. 

Connected morphological operations  initially proposed by Salambier and Serra \cite{salembier1995flat,serra1993connected}, has been  investigated extensively by many researchers \cite{heijmans1999connected}, \cite{ronse1998set}, \cite{serra1998connectivity}. These kind of operators act on gray-level images owing to the concept of flat zones. Simply in binary case, a connected operator is not capable of modifying background or foreground boundaries of an image. Roughly speaking, connected morphological operators interact with a binary image just by preserving or removing connected components. Figure 2 illustrates an example of a connected and non-connected operator. A connected morphological operation can't produce the image in the middle since the boundaries of foreground image is moved. Precisely, a connected morphological operator is just capable of disappearing or creating connected components of an image but not moving or breaking any of their boundaries.

However, eroding a graph by another structuring graph may not be classified as a connected operation. Figure 4 illustrates an example of eroding a graph by a structuring graph formed from a triangle consisting of a root and two buds which is not a connected operation. In fact many practical properties of dilation and erosion do not hold in the graph area. One of them is the anti-extensive property of erosion which is witnessed to be violated in figure 4. The reason for extending the foreground stems from the definition of erosion. The erosion of graph $X$ representing a foreground image by the structuring graph $S$ is calculated via: 
\begin{equation}
\epsilon_S(X|G)=\{v \in V|N_A(v|G) \subseteq X \}
\label{eq:1}
\end{equation}
$X,G$ stands for the graphs representing foreground and whole image and $N_A(v|G)$ is the neighborhood function defined by the formula: 
\begin{multline}
N_S(v|G)=\\ \Union{\theta}{\theta \mbox{ is an embedding of $S$ into $G$ at node $v$}}{\theta(B_S)}
\end{multline} 
Roughly speaking, for deciding whether a node $v$ of the  graph presenting a picture will be black or white after the erosion operation, one looks for all embeddings of the structuring graph in the source graph in which a root of the structuring graph is mapped to $v$. If buds of the structuring element are all mapped to nodes with black color (foreground graph), then $v$ becomes black, otherwise white. However, in the case in which not any  embedding of the structuring graph into the source exists at node $v$ exists, $N_A(v|G)$ is empty and $v$ becomes black since $\emptyset \subseteq X$ always holds. This confirms that the foreground pixels(black nodes) may be extended via an erosion operation and the erosion may extend the boundaries of the foreground part which contradicts with the definition of connected morphological operations.

It is obvious that the aforementioned erosion is not connected since it changes the boundaries of foreground hexagon and this pretty assures that the skeletonization of a graph by an structuring graph may also be non-connected.

However, one can deduce  that deciding about satisfying the connectivity property for structured graph morphological operations can be quite a difficult problem. Deciding on the problem that the structuring graph can be embedded in the source graph reduces to the subgraph isomorphism problem which is itself the generalization of maximum clique problem and is proved to be NP-complete \cite{cook1971complexity}. Investigating the conditions for connectivity of morphological operations like dilation and erosion for non-structured graph morphology looks more interesting since we are not facing the problem of finding the embeddings of the structured graph into the source graph. This is an interesting future research topic.

\begin{figure*}[h]
	\centering
	\includegraphics[scale=0.65]{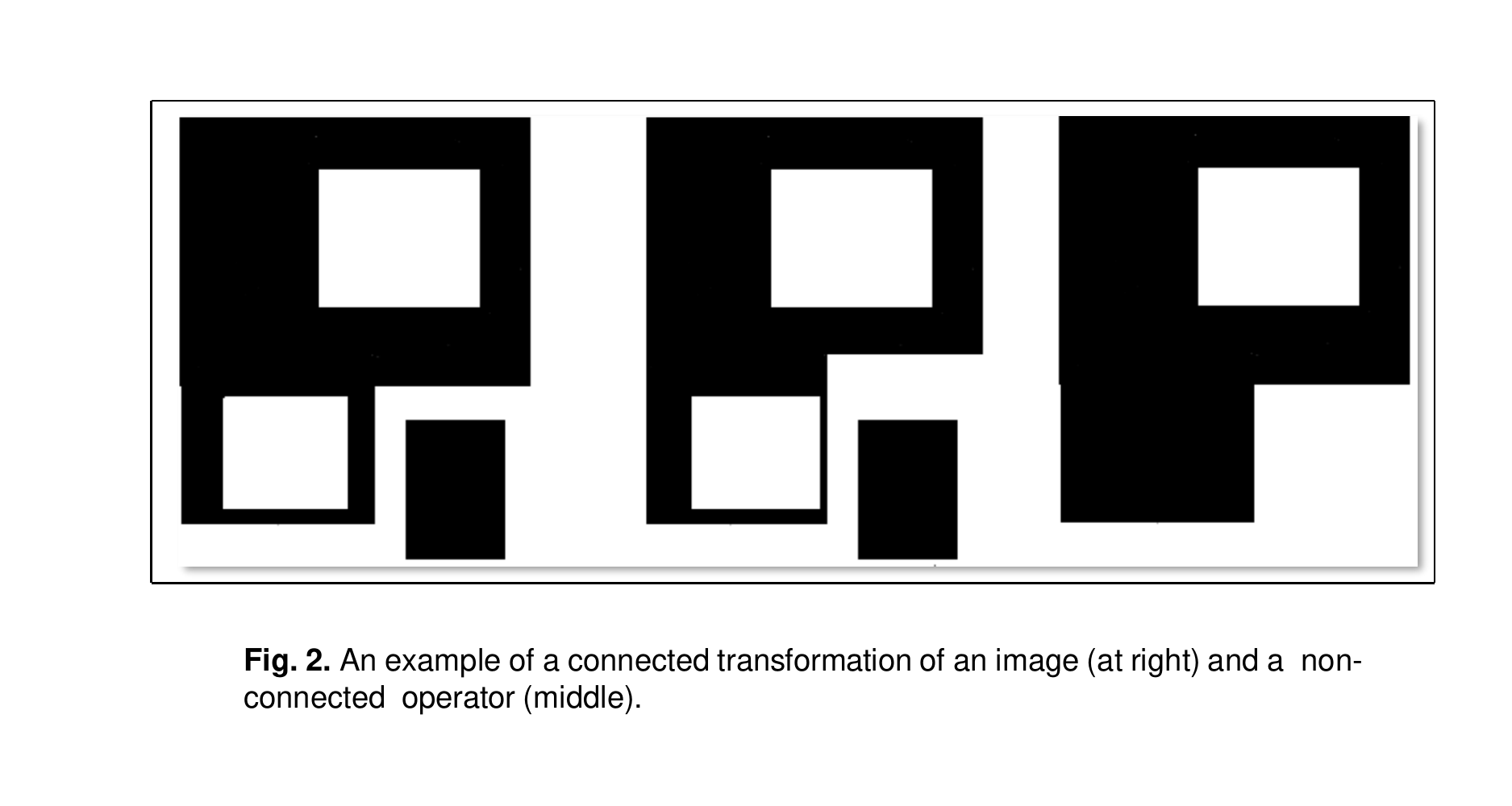}
	\label{figure:1}
\end{figure*}

\section{Method}
\label{sec:examples}
Given a 4,8-connected grid represented by an undirected graph $G=(V_G,E_G)$, suppose an image is represented by $M=(V_M,E_M)$ satisfying $V_M \subseteq V_G, E_M \subseteq E_G$. The inclusion is so trivial since the image is embedded in the grid. We observe just the binary graphs and the method which will be proposed corresponds to this category.

Inspiring from the vertex based structured graph morphology introduced in \cite{heumans1992graph} and the non-structured edge based morphological operations proposed in \cite{cousty2013morphological}, we came to the structured edge based morphological operations. We do not include any explanation on aforementioned methods for the sake of brevity but the reader can refer to aforementioned papers. The idea of homomorphisms from the structuring graph to the source graph is the core of node based structured graph morphological transformations. Nodes of the structuring graph are classified to 3 categories called roots, buds and nodes that are neither roots nor buds. If an injective homomorphism from the structuring graph to the source graph exists in which a root node is mapped to a node $v$ with value 1 and the mapping of at least one bud is a node with the value 1, then $v$ is added to the result of dilation operation. 
Inspiring from this idea, our approach is constructed on injective homomorphisms from the structuring graphs to the source graph as well. The edges of structuring graph are classified as roots, buds, not root, bud edges.

Imagine an edge $e$ in the source graph. If an injective homomorphism from the structuring graph to the source graph exists that maps a root edge  to  $e$,  then the mapping of  buds are added to the result of dilation operation. This technique is axiomatized as follows.

Given a structuring graph $S=(V_S,E_S)$, define two sets $R
_S,B_S \subseteq E_S$, as the set of roots and buds respectively. Suppose the regular graph denoting the grid as $G$ and the graph that represents the image is denoted by $M$. If $\theta:V_S \to V_G$  is a homomorphism and $h:E_S \to E_G$ is defined for any $(u,v) \in E_S$, as $h(u,v)= (\theta(u),\theta(v))$.  The neighborhood function of an edge $e$ can now be defined by,
\begin{equation}
N_S(e|G)=\Union{h}{h \mbox{ maps an edge in } R_S \mbox{ to } e}{h(B_S)}
\label{fig5}
\end{equation}
In fact this corresponds to the informal description of dilation. if there exist a homomorphism $h$ which maps an edge  of the structuring element defined as root to an edge $e$ of the foreground image, the mappings of edges defined as buds are along with the vertices coinciding with them are added to the result. The neighborhood function $N_S(e|G)$  extracts the mapping of buds regarding any homomorphism $h$. Now the dilation of graph G by S as the structuring element is defined by,
\begin{equation}
\delta(M|G)=(\Union{e}{e \in M}{N_S(e|G)}.
\label{eq:4}
\end{equation}

  \begin{figure*}[!b]
 	\centering
 	\includegraphics[scale=0.37]{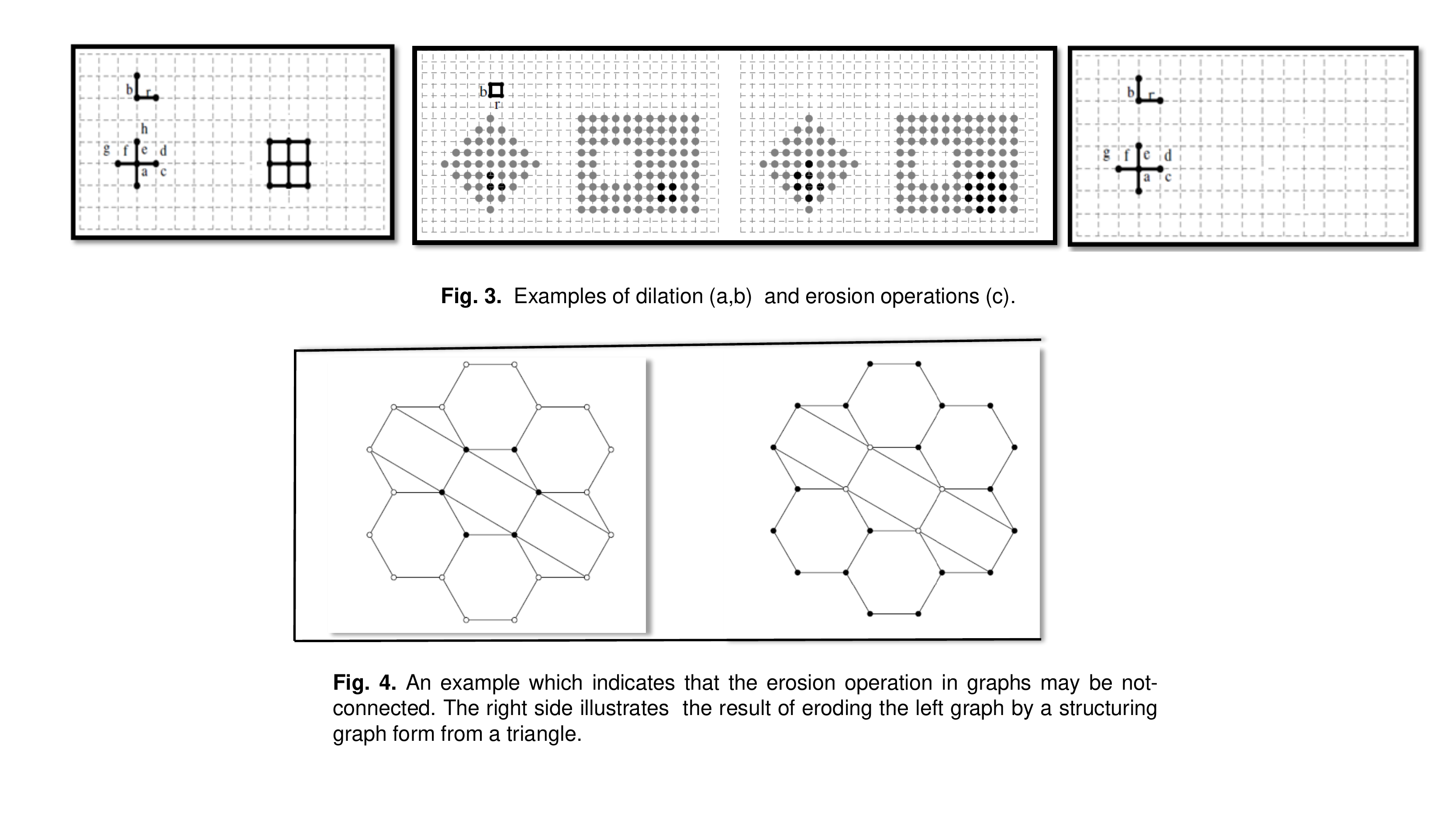}
 	\label{fig:5}
 \end{figure*}
Figure 3 illustrates an example of dilation transformation. It should be noted that the graph dilation operation is denoted by $\delta$ and the erosion by $\epsilon$ and $G$ corresponds to the graph shown by dashed line. The structuring graph consists of 3 nodes and 2 edges with the only root edge labeled by 'r' and the only bud labeled by 'b'. For instance assume the homomorphism $h:r \to (a,c)$. This embedding results in adding $(d,c)$, since it is the mapping of the only bud. Mapping the root edge of the structuring graph to $(a,e)$ entails adding $(d,e)$ to the result of dilation according to \ref{eq:4}.
Another example illustrating dilation of an image composed of two disjoint graphs taking a square structuring element including one root edge and one bud edge is also shown in figure 3(b). 
Getting the dilation in hand, the erosion can be calculated making use of its Galois connection with the dilation.
\begin{definition} \cite{backhouse1998pair}
Given sets $P, Q$ and their powersets $\rho(P),\rho(Q)$, a pair of functions $f:\rho(P) \to \rho (Q), g: \rho(Q) \to \rho (P)$
constitute a Galois connection if for each  $X \in \rho(P) \wedge Y \in \rho(Q)$ we have: \begin{center} $X \subseteq f(y)  \iff Y \subseteq g(X)$
	\end{center}
\end{definition}
The function $f$ is called the lower adjoint of $g$ and $g$  the upper adjoint of $f$. A very useful property of a Galois connection is that an upper or lower adjoint function uniquely determines the other saying that $f(a)$ is the least $d$ satisfying $a \leq g(d)$ and $g(b)$ is the largest existent $c$ where $f(c) \leq d$ \cite{ore1944galois}.
It is pretty known that dilation and erosion engage in a Galois connection \cite{heijmans1990algebraic}. This allows to extract the axiomatization of erosion from dilation.
According to the Galois connection between dilation and erosion we can write: \begin{center} $\delta_S(M|G) \subseteq P \iff M \subseteq \epsilon_S(P|G)$. \end{center}
\begin{deriv}
	\delta_S(M|G) \subseteq P
	\<\ssi
	\commentaire{Definition of dilation}
	\Union{e}{e \in E_M}{N_S(e|G) \subseteq P} 
	\<\ssi
	\commentaire{Properties of $\cup$}
	\pourtout{e}{e \in E_M}{N_S(e|G) \subseteq P}	
	 \<\ssi
	 \commentaire{$E_M \subseteq E_G$}
	 G \subseteq \pourtout{e}{e \in E_G}{N_S(e|G) \subseteq P} 
	\<\ssi
	\commentaire{Properties of $\forall$} 
	G \subseteq \Inter{e}{e \in E_G}{N_S(e|G) \subseteq P}
\end{deriv}
	This yields that the erosion can be calculated by: 
	\begin{equation}
	\label{eq:2}
	\epsilon_S(P|G)=\Inter{e}{e \in E_G}{N_S(e|G) \subseteq P}.
	\end{equation}
Equation \ref{eq:2} provides a procedure for calculating the erosion of a graph $G$ by means of the structuring element $S$. Imagine $G$ represents the 4 or 8-connected grid that corresponds to the whole 2-dimensional plane and $P$ is the graph representation of the image. If a root edge of the structuring element is mapped to an edge $e$ in $P$ while all edges of $G$ that are neighbors of mappings of buds are included in $P$, $e$ is added to the result of erosion. 

Figure 3(c) provides an example of the erosion operation. $G$ corresponds to the whole plane (dashed lines) and $P$ is indicated in the left. The result of erosion operation is empty. Imagine the embedding $h:r \to (a,e)$. Since $(e,d)$ is not an edge of the graph, nothing is added to the result of erosion. So, If the mapping of buds through all existent embeddings are contained in the source graph, the edges which are the mapping of roots are added to the result of erosion. 

 
With dilation and erosion in hand, one can simply express the opening and closing. However, there is another way of specifying opening and closing called structural opening \cite{heumans1992graph}.
One can also define the opening transform for structured morphological graph transformations based on edges as follows: $\alpha_S(M|G)= \Union{h}{h \mbox { is a homomorphism and   } h(B_S) \subseteq M}{h(B_S)}$.
This definition states that the opening consists of buds mappings whenever mappings of all buds is included in the source graph. It is not hard to prove that this formula is indeed an opening. For that target we need to prove that $\alpha_S(M|G)$ is anti-extensive, idempotent and increasing.
\begin{definition}
 Let $O: X \to X $ be a function acting on complete lattice $X$. One  can express that:
\begin{itemize}
\item  $O$ is idempotent if $O^2=O$.
\item $O$ is anti-extensive if for every $u \in X$ then $O(u) \leq u$.
\item $O$ is increasing if for every $u, v \in X$,$ u \leq v \imp O(u) \leq O(v)$.
\end{itemize}
Proving the anti-extensive property is straightforward since $h(B_S) \subseteq M \imp \alpha_S(M|G) \subseteq M$. The proof for increasing property is that $M_1 \subseteq M_2 \imp \alpha_S(M_1|G) \subseteq \alpha_S(M_2|G)$ because $M_1 \subseteq M_2$ and $ h(B_S)) \subseteq M_1$ for a homomorphism h implies that $h(B_S) \subseteq M_2$ . For proving idempotency we have $ \alpha_S(\alpha_S(M|G)) \subseteq \alpha_S(M|G)$ by the anti-extensive property. For any edge $e \in \alpha_S(M|G)$, we have $e \in \alpha_S(\alpha_S(M|G))$. Thus $\alpha_S(\alpha_S(M|G))= \alpha_S(M|G)$ .
\end{definition}
The primary procedure for extracting the skeleton of an image requires dilating the structuring element by itself for $n$ times and then eroding the source image by this structuring graph. However,one can erode the image by the structuring element $n$ times instead. Algorithm \ref{alg:skel} narrates the procedure performed for extracting the skeleton of an image formed as a graph $M$ with a predefined structuring graph $S$. 
There $\epsilon_S(M|G)$ denotes eroding the picture represented by graph $M$ with the structuring graph $S$ and $ \delta_S(M|G)^{n+1}$ stands for dilating graph $M$ by $S$. Instead of dilating the structuring graph $n$ times with itself, the algorithm performs the dilation with the structuring graph $n$ times which is the common way of computing the skeleton in literature. In fact The algorithm formalizes the following formula for computing the skeleton of an image shown with the graph $M$ which is embedded in the grid indicated by the regular graph $G$.

\begin{equation}
\mathsf{skel}_n(M)=\epsilon_S(M|G)^n-\delta_S(\epsilon_S(M|G)^{n+1})
\end{equation}
 
\begin{algorithm}[!t]
	\caption{Skeletonization algorithm }\label{alg:skel}
	\begin{algorithmic}[1]
		\Procedure{Skel}{$G,M,S$}\Comment{$S$ structuring -graph, $M$ source and $G$ the grid}
		\State $S \gets \emptyset, Temp \gets \delta_\emptyset(M|G)$
		\While{$temp \not= \emptyset$}
		\State $Temp \gets \epsilon_S(M|G)$ \Comment{erosion-$\epsilon_S(M|G)^n$}
		\State $Temp_1 \gets \epsilon_S(Temp|G)$ \Comment{erosion once more -$\epsilon_S(M|G)^{n+1}$}
		\State $Temp_2 \gets \delta_S(Temp_1|G)$ \Comment{dilation-$\delta_S(\epsilon_S(M|G)^{n+1})$}
		\State $S \gets S \cup Temp_2$
		\EndWhile\label{euclidendwhile}
		\State \textbf{return  $Temp_2$ } \Comment{The skeleton is $Temp_2$}
		\EndProcedure
	\end{algorithmic}
\end{algorithm}
Figure 5 indicates an example of skeletonization. On contrary with the skeleton algorithm for showing what happens more clearly, instead of expressed method performing $n$ consecutive erosions,  first the structuring graph is dilated by itself in $n$ times corresponding to the initiative approach. Intuitively, the role of disk with radius $n$ is simulated by regular graphs. 


Conversely, We can reconstruct a graph M from its skeleton by the following formula:
\begin{equation}
\label{eq3}
\delta_S^k \epsilon_S^k= \Union{n}{k \leq n \leq N }{\delta_S^n(\mathsf{skel}_n(M))}.
\end{equation} 
Specifying k=0 yields almost the exact reconstruction of the source image. The benefit of reconstruction procedure is that all subsets of the skeleton derived in each iteration of the skeletonization algorithm resembles the original image and the original image can be reconstructed from them. However, if $k>0$ then the original image can be partially reconstructed and an opened copy of the original shape is rebuilt according to \ref{eq3}. The algorithm \ref{alg:rec} states the procedure formally. 
\begin{algorithm}[!t]
		\caption{Reconstruction algorithm }
	\begin{algorithmic}[1]
		\Procedure{Recons}{$G,k,M_k,S$}\Comment{$S$ structuring -graph, $M_k$ the result of skel algorithm until step $k$ and $G$ the grid}
		\State $ Count=k,Temp=M_k$
		\While{$count  \leq N$}
		 
		\State $Temp \gets \delta_S(Temp|G)$ \Comment{dilation-$\delta_S(M_k|G)$}
				\State $Count \gets Count+1$
				\EndWhile\label{euclidendwhile1}
		\State \textbf{return  $Temp$ } \Comment{The skeleton is $Temp$}
		\EndProcedure
	\end{algorithmic}
\label{alg:rec}
\end{algorithm}

An advantage of skeletonization and reconstruction built on 
edge based morphological operations comparing to the vertex based method is that one can import quite rich and applicative concept of graph paths into it. A path is a sequence of vertices in a graph in which every two successive vertices are connected by an edge. If weights are imposed to edges, the length of a path is considered to be the sum of edge weights constituting the path. Otherwise, for an unweighted graph the  length of any path is simply carried out by counting the number of its consisting edges. 

An important concept directly deduced from paths which is widely utilized in many applications is the distance map. The distance map between two pixels $p$ and $q$ simply is the length  of shortest path connecting them. The connecting path between any two pixels is calculated regarding to the connectivity policy which can be 4,6,8-connected. A well-known method of computing the skeleton of a binary image \cite{di1994well} , \cite {arcelli1993euclidean} relies on the distance map of foreground pixels calculated from the closest background pixel.

\begin{figure*}[htbp]
	\centering
	\includegraphics[scale=0.47]{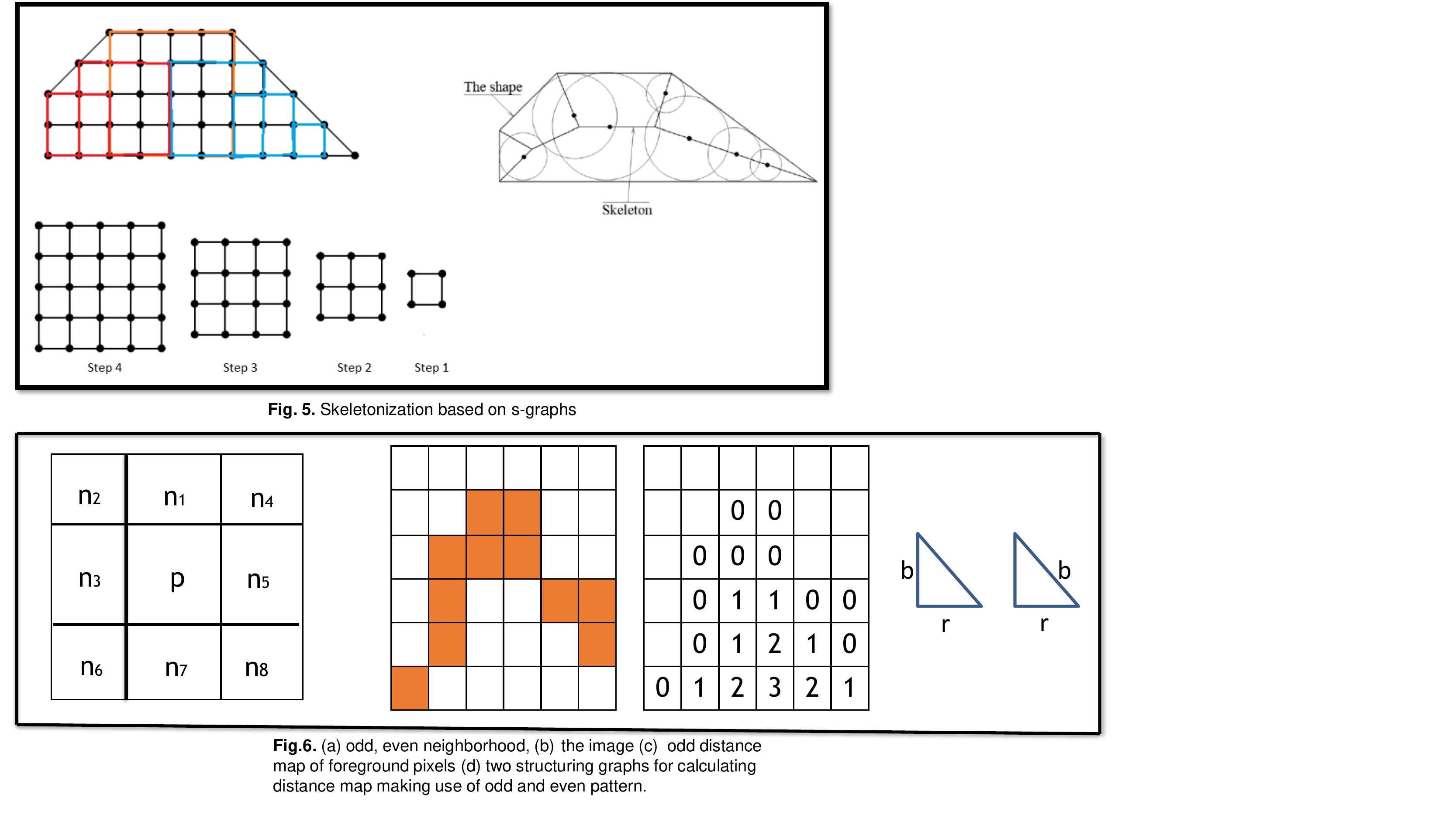}
	\label{fig:8}
\end{figure*}

\begin{figure*}[!t]
	\centering
	\includegraphics[scale=0.35]{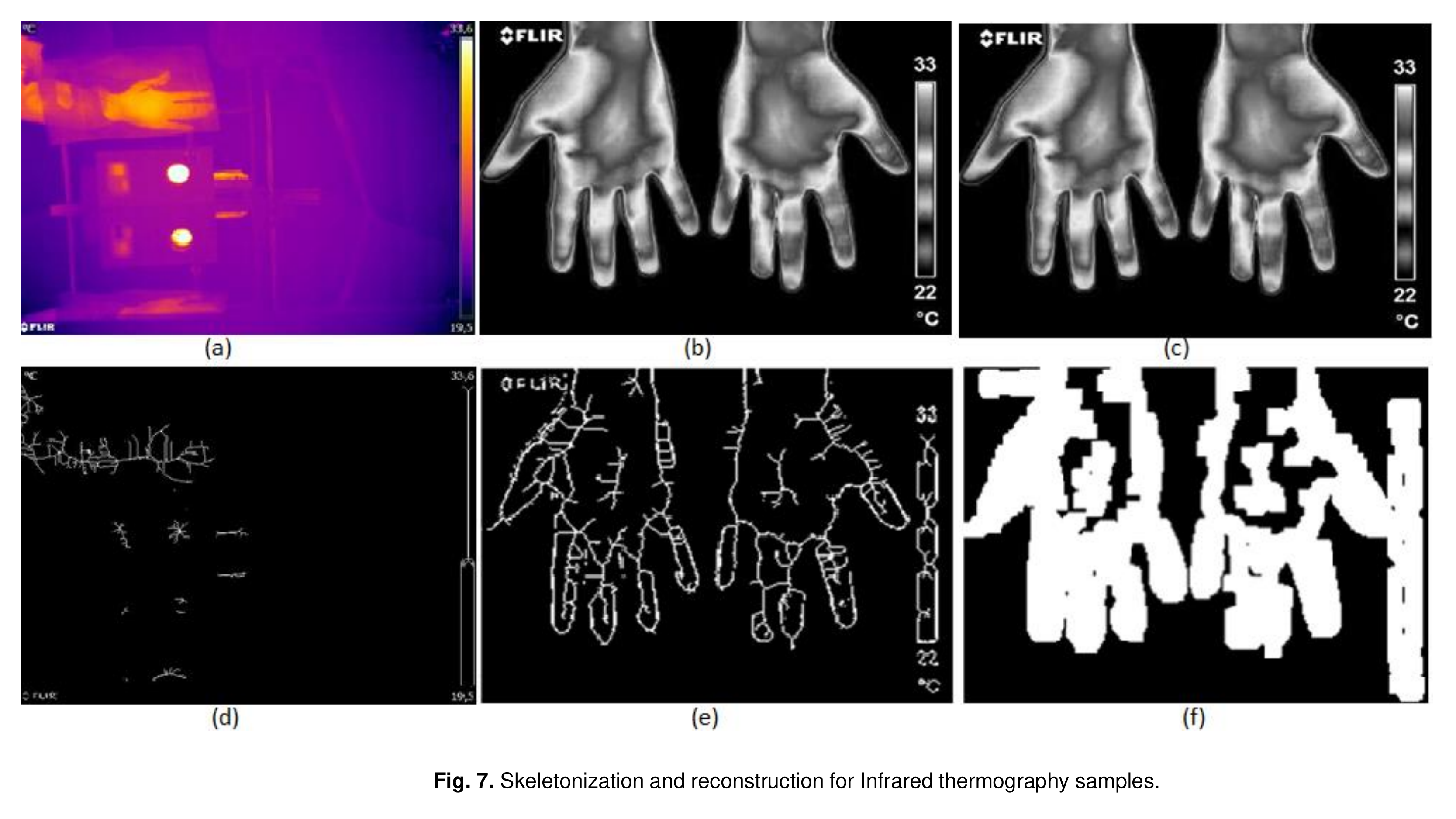}
	\label{fig:7}
\end{figure*}

A brief description of skeletonization of a binary image emerged from distance maps is expressed as follows. Refer to  \cite{di1994well} for the detailed description of the method. The distance map between two pixels $p$ and $m$ is computed either by $D_\mathsf{odd}$ or $D_\mathsf{even}$ depending on choosing horizontal/vertical or diagonal steps for the path between them. Figure 6 illustrates $D_\mathsf{odd}$ or $D_\mathsf{even}$. Intuitively, $D_\mathsf{odd}$ is denoted by $D_{1,3,5,7}$ while $D_\mathsf{even}$ is written as $D_{2,4,6,8}$. The distance map of pixels in each foreground sections of the image from the closest pixel on the boundary are calculated using either the $D_\mathsf{odd}$ or $D_\mathsf{even}$ pattern. The local extremes of the distance maps corresponding to the image foreground pixels constitute the skeleton. Figure 6d illustrates a $D_\mathsf{odd}$ distance maps of the foreground pixels from the boundary pixels. The pixels with maximal distance maps saying with distance maps  2,3 constitute the skeleton of the image. This distance maps can be calculated by edge based erosion operations. In other words, continuing the edge based erosion of the source graph with the structuring graph until the two pixels $p$ and $m$ does not both include in the result provides the distance map between $p$ and $m$ by the number of erosions. Different connectivity policies for calculation of distance maps can be easily covered by selecting different edges of the structuring graph as roots and buds. For instance, one can use the first triangle shown in figure 6(d) for calculating the distance map $D_\mathsf{odd}$ whereas the distance map $D_\mathsf{even}$  can be calculated making use of the second triangle as the structuring element. If an embedding  maps the root in the left triangle to an edge that contains a pixel $p$, the pixels with the horizontal/vertical distance map 1 from $p$ are contained in the erosion of image with that triangle. Similarly, If an embedding  maps the root in the right triangle to an edge that contains a pixel $p$, the pixels with the diagonal distance map 1 from $p$ are contained in the erosion of image with that triangle.

Let us express IFT as  another interesting application of edge based morphological operations that is out of reach with the vertex based method. Image foresting transform (IFT) proposed by Falcao et al. \cite{falcao2004image} reminds a widely used optimal path algorithm in image processing. IFT intuitively suggests searching for minimal paths to pixels from other pixels. In graph based representation of images this resembles the well-known idea of minimum spanning tree but with the difference that the optimal paths needs not to be connected but forming a forest. The idea for skeletonization of a gray-level image nominated in \cite{falcao2004image} suggests making use of minimal paths from pixels on the boundary of a foreground image to the pixels of foreground for computing its skeleton. This idea can be implemented by our edge based morphological transformations. General definition of erosion and dilation for gray-level images are required for this purpose which are given  respectively,
\begin{equation}
N_S(e|G)=\Max{h}{h \mbox{ maps an edge in } R_S \mbox{ to } e}{h(B_S)}
\label{fig8}
\end{equation}
\begin{equation}
\epsilon_S(P|G)=\Min{e}{e \in E_G}{N_S(e|G) \subseteq P}.
\label{eq5}
\end{equation}
In simple words in dilation the weight of each edge changes to maximum weight of buds mapped to it through different homomorphisms($h$). Similarly, within the erosion the weight of each edge changes to minimum weight of buds mapped to it through different homomorphisms($h$). Now, one may reformulate the skeletonization algorithm 1 for gray-scaled images substituting the new definition of edge based dilation and erosion \ref{fig8}, \ref{eq5} for erosion and dilation operations. IFT calls the pixels that are used as the source of minimal paths to a pixel seeds. For calculating the skeleton of an image, it calculates the minimal weight paths from pixels on the boundary to the image pixels. The minimal path from a pixel $p$ to a pixel $q$ can be extracted by iterative gray-scaled erosions containing $p$, $q$. For the same reason triangular structuring graphs with different root and buds can collaborate the erosion with one of 4,8-connected policies.  

To summarize, morphological operations of a graph with some structuring graph based on edges opens the door to use lots of ideas relevant to optimal maps that far from our access within the vertex based morphological operations.

\section{Results}
We have applied the expressed approach in order to to obtain the skeleton of some infrared thermal images. Since the implementation of the method for grayscale images remains for future research, the sample infrared images were transformed to binary counterparts and then their skeleton was calculated. Figure 7(a) illustrates the sample image . Its skeleton using the structuring graph isomorphic to 3*3 grid is depicted in 7(d). Figure 7(e) represents the skeleton of 7(b) by the same structuring graph.  Figure 7(f) illustrates the result of reconstruction of the source image using the skeleton 7(e). The value $k=0$ was employed along with the reconstruction operation aiming at reconstructing the exact original image. The partial reconstruction of the original image by $k=1$ was not so satisfactory.

\section{Conclusion}
A new method for extracting skeleton of a binary image using graph morphological operations based on edges have been proposed. This approach enables us to import many widespread path based methods into morphological graph operations which are not accessible with the well-known vertex based methods. Some experiments have been conducted using this approach aiming at extracting the skeleton of infrared thermography images and reconstructing the original image from the skeleton. Investigating on edge based implementation of various path based techniques can be a potential research subject.

\section{Funding Information}
The research was supported by the Natural Science and Engineering Research Council of Canada (NSERC). The experiments of this research were conducted under Canada research chair in Multipolar Infrared Vision (MIVIM).

\bibliographystyle{IEEEtran}
\bibliography{sample}

\begin{thebibliography}{10}
\providecommand{\url}[1]{#1}
\csname url@samestyle\endcsname
\providecommand{\newblock}{\relax}
\providecommand{\bibinfo}[2]{#2}
\providecommand{\BIBentrySTDinterwordspacing}{\spaceskip=0pt\relax}
\providecommand{\BIBentryALTinterwordstretchfactor}{4}
\providecommand{\BIBentryALTinterwordspacing}{\spaceskip=\fontdimen2\font plus
\BIBentryALTinterwordstretchfactor\fontdimen3\font minus
  \fontdimen4\font\relax}
\providecommand{\BIBforeignlanguage}[2]{{%
\expandafter\ifx\csname l@#1\endcsname\relax
\typeout{** WARNING: IEEEtran.bst: No hyphenation pattern has been}%
\typeout{** loaded for the language `#1'. Using the pattern for}%
\typeout{** the default language instead.}%
\else
\language=\csname l@#1\endcsname
\fi
#2}}
\providecommand{\BIBdecl}{\relax}
\BIBdecl

\bibitem{heijmans1990algebraic}
H.~J. Heijmans and C.~Ronse, ``The algebraic basis of mathematical morphology
  i. dilations and erosions.''\hskip 1em plus 0.5em minus 0.4em\relax Elsevier,
  1990, vol.~50, no.~3, pp. 245--295.

\bibitem{x1}
Y.~B. Sharifipour~HM, ``Mathematical morphology via category theory,''
  \emph{arXiv preprint: arXiv:2009.06127}, 2020.

\bibitem{x2}
H.~Memarzadeh~sharifipour, B.~Yousefi, and X.~Maldague, ``Skeletonization and
  reconstruction based on graph morphological transformations,'' \emph{Advanced
  Infrared Technology and Applications Conference}, 2017.

\bibitem{x3}
B.~Yousefi, S.~Sfarra, C.~I. Castanedo, and X.~P. Maldague, ``Comparative
  analysis on thermal non-destructive testing imagery applying candid
  covariance-free incremental principal component thermography (ccipct),''
  \emph{Infrared Physics \& Technology}, vol.~85, pp. 163--169, 2017.

\bibitem{x4}
B.~Yousefi, C.~I. Castanedo, X.~P. Maldague, and G.~Beaudoin, ``Assessing the
  reliability of an automated system for mineral identification using lwir
  hyperspectral infrared imagery,'' \emph{Minerals Engineering}, vol. 155, p.
  106409, 2020.

\bibitem{x5}
B.~Yousefi, H.~Memarzadeh~Sharifipour, M.~Eskandari, C.~Ibarra-Castanedo,
  D.~Laurendeau, R.~Watts, M.~Klein, and X.~P. Maldague, ``Incremental low rank
  noise reduction for robust infrared tracking of body temperature during
  medical imaging,'' \emph{Electronics}, vol.~8, no.~11, p. 1301, 2019.

\bibitem{x6}
B.~Yousefi, S.~Sfarra, F.~Sarasini, C.~I. Castanedo, and X.~P. Maldague,
  ``Low-rank sparse principal component thermography (sparse-pct): Comparative
  assessment on detection of subsurface defects,'' \emph{Infrared Physics \&
  Technology}, vol.~98, pp. 278--284, 2019.

\bibitem{yousefi2014hierarchical}
B.~Yousefi, S.~M. Mirhassani, A.~AhmadiFard, and M.~Hosseini, ``Hierarchical
  segmentation of urban satellite imagery.''\hskip 1em plus 0.5em minus
  0.4em\relax Elsevier, 2014, vol.~30, pp. 158--166.

\bibitem{yan2015infrared}
X.~Yan, H.~Qin, J.~Li, H.~Zhou, and J.-g. Zong, ``Infrared and visible image
  fusion with spectral graph wavelet transform.''\hskip 1em plus 0.5em minus
  0.4em\relax Optical Society of America, 2015, vol.~32, no.~9, pp. 1643--1652.

\bibitem{vollmer2017infrared}
M.~Vollmer and K.-P. M{\"o}llmann, \emph{Infrared thermal imaging:
  fundamentals, research and applications}.\hskip 1em plus 0.5em minus
  0.4em\relax John Wiley \& Sons, 2017.

\bibitem{elibol2014graph}
A.~Elibol, N.~Gracias, R.~Garcia, and J.~Kim, ``Graph theory approach for match
  reduction in image mosaicing,'' \emph{JOSA A}, vol.~31, no.~4, pp. 773--782,
  2014.

\bibitem{biumtransformation}
H.~Bium, ``A transformation for extracting new descriptions of shape,'' in
  \emph{Symposium on Modeis for the Perception of Speech and Visua1 Form}.

\bibitem{beucher1994digital}
S.~Beucher, ``Digital skeletons in euclidean and geodesic spaces.''\hskip 1em
  plus 0.5em minus 0.4em\relax Elsevier, 1994, pp. 127--141.

\bibitem{lantuejoul1978squelettisation}
C.~Lantuejoul, ``La squelettisation et son application aux mesures topologiques
  des mosaiques polycristallines, th{\`e}ses dr,'' 1978.

\bibitem{kresch1994morphological}
R.~Kresch and D.~Malah, ``Morphological reduction of skeleton
  redundancy.''\hskip 1em plus 0.5em minus 0.4em\relax Elsevier, 1994, vol.~38,
  no.~1, pp. 143--151.

\bibitem{maragos1986morphological}
P.~Maragos and R.~Schafer, ``Morphological skeleton representation and coding
  of binary images.''\hskip 1em plus 0.5em minus 0.4em\relax IEEE, 1986,
  vol.~34, no.~5, pp. 1228--1244.

\bibitem{maragos2013segmentation}
P.~Maragos and K.~Drakopoulos, ``Segmentation and skeletonization on arbitrary
  graphs using multiscale morphology and active contours,'' in
  \emph{Innovations for Shape Analysis}.\hskip 1em plus 0.5em minus 0.4em\relax
  Springer, 2013, pp. 53--75.

\bibitem{maragos1990morphological}
P.~Maragos and R.~W. Schafer, ``Morphological systems for multidimensional
  signal processing.''\hskip 1em plus 0.5em minus 0.4em\relax IEEE, 1990,
  vol.~78, no.~4, pp. 690--710.

\bibitem{vincent1989graphs}
L.~Vincent, ``Graphs and mathematical morphology.''\hskip 1em plus 0.5em minus
  0.4em\relax Elsevier, 1989, vol.~16, no.~4, pp. 365--388.

\bibitem{najman2012short}
L.~Najman and F.~Meyer, ``A short tour of mathematical morphology on edge and
  vertex weighted graphs.''\hskip 1em plus 0.5em minus 0.4em\relax CRC Press,
  2012.

\bibitem{salembier1995flat}
P.~Salembier and J.~Serra, ``Flat zones filtering, connected operators, and
  filters by reconstruction,'' \emph{IEEE Transactions on image processing},
  vol.~4, no.~8, pp. 1153--1160, 1995.

\bibitem{serra1993connected}
J.~C. Serra and P.~Salembier, ``Connected operators and pyramids,'' in
  \emph{Image algebra and morphological image processing IV}, vol. 2030.\hskip
  1em plus 0.5em minus 0.4em\relax International Society for Optics and
  Photonics, 1993, pp. 65--77.

\bibitem{heijmans1999connected}
H.~J. Heijmans, ``Connected morphological operators for binary images,''
  \emph{Computer Vision and Image Understanding}, vol.~73, no.~1, pp. 99--120,
  1999.

\bibitem{ronse1998set}
C.~Ronse, ``Set-theoretical algebraic approaches to connectivity in continuous
  or digital spaces,'' \emph{Journal of Mathematical Imaging and Vision},
  vol.~8, no.~1, pp. 41--58, 1998.

\bibitem{serra1998connectivity}
J.~Serra, ``Connectivity on complete lattices,'' \emph{Journal of Mathematical
  Imaging and Vision}, vol.~9, no.~3, pp. 231--251, 1998.

\bibitem{cook1971complexity}
S.~A. Cook, ``The complexity of theorem-proving procedures,'' in
  \emph{Proceedings of the third annual ACM symposium on Theory of
  computing}.\hskip 1em plus 0.5em minus 0.4em\relax ACM, 1971, pp. 151--158.

\bibitem{heumans1992graph}
H.~Heumans, P.~Nacken, A.~Toet, and L.~Vincent, ``Graph morphology.''\hskip 1em
  plus 0.5em minus 0.4em\relax Elsevier, 1992, vol.~3, no.~1, pp. 24--38.

\bibitem{cousty2013morphological}
J.~Cousty, L.~Najman, F.~Dias, and J.~Serra, ``Morphological filtering on
  graphs.''\hskip 1em plus 0.5em minus 0.4em\relax Elsevier, 2013, vol. 117,
  no.~4, pp. 370--385.

\bibitem{backhouse1998pair}
R.~Backhouse, ``Pair algebras and galois connections.''\hskip 1em plus 0.5em
  minus 0.4em\relax Elsevier, 1998, vol.~67, no.~4, pp. 169--175.

\bibitem{ore1944galois}
O.~Ore, ``Galois connexions.''\hskip 1em plus 0.5em minus 0.4em\relax JSTOR,
  1944, vol.~55, no.~3, pp. 493--513.

\bibitem{di1994well}
G.~S. di~Baja, ``Well-shaped, stable, and reversible skeletons from the (3,
  4)-distance transform,'' \emph{Journal of visual communication and image
  representation}, vol.~5, no.~1, pp. 107--115, 1994.

\bibitem{arcelli1993euclidean}
C.~Arcelli and G.~S. di~Baja, ``Euclidean skeleton via centre-of-maximal-disc
  extraction,'' \emph{Image and Vision Computing}, vol.~11, no.~3, pp.
  163--173, 1993.

\bibitem{falcao2004image}
A.~X. Falc{\~a}o, J.~Stolfi, and R.~de~Alencar~Lotufo, ``The image foresting
  transform: Theory, algorithms, and applications,'' \emph{IEEE transactions on
  pattern analysis and machine intelligence}, vol.~26, no.~1, pp. 19--29, 2004.

\end{thebibliography}

\end{document}